\documentclass{article}
\usepackage{algorithm}
\usepackage{algorithmic}
\usepackage{graphicx} 
\usepackage{mlapa}
\usepackage{amssymb}

\newtheorem{thm}{Theorem}[section]

\newtheorem{lem}[thm]{Lemma}

\newtheorem{rem}[thm]{Remark}

\newcommand{\Real}{\mathbb R}

\newcommand{\rmax}{R_{\max}}
\newcommand{\vmax}{V_{\max}}

\newcommand{\ourmethod}{{\sc{OIM}}}
\newcommand{\vo}{\vmax}
\newcommand{\ro}{\rmax}
\newcommand{\xsize}{\left|X\right|}

\newcommand{\citep}[1]{\cite{#1}}
\newcommand{\citet}[1]{\emcite{#1}}

\newcommand{\bproof}{\textit{Proof. }}
\newcommand{\eproof}{\hfill $\blacksquare$}

\begin{document}

\title{The many faces of optimism \\ Extended version}


 \author{Istv{\'a}n Szita \and Andr{\'a}s L{\H{o}}rincz}

\maketitle

\begin{abstract}
The exploration-exploitation dilemma has been an intriguing and
unsolved problem within the framework of reinforcement learning.
``Optimism in the face of uncertainty'' and model building play
central roles in advanced exploration methods. Here, we integrate
several concepts and obtain a fast and simple algorithm. We show
that the proposed algorithm finds a near-optimal policy in
polynomial time, and give experimental evidence that it is robust
and efficient compared to its ascendants.
%
\end{abstract}


\section{Introduction}

Reinforcement learning (RL) is the art of maximizing long-term
rewards in a stochastic, unknown environment. In the construction
of RL algorithms, the choice of exploration strategy is of central
significance.

We shall examine the problem of exploration in the Markov decision
process (MDP) framework. While simple methods like
$\epsilon$-greedy and Boltzmann exploration are commonly used, it
is known that their behavior can be extremely poor
\citep{Koenig93Complexity}. Recently, a number of efficient
exploration algorithms have been published, and for some of them,
formal proofs of efficiency also exist. We review these methods in
Section~\ref{s:literature}. By combining ideas from several
sources, we construct a new algorithm for efficient exploration.
The new algorithm, \emph{optimistic initial model} (\ourmethod),
is described in Section~\ref{s:algorithm}. In
Section~\ref{s:analysis}, we show that many of the advanced
algorithms, including ours, can be treated in a unified way. We
use this fact to sketch a proof that \ourmethod\ finds a
near-optimal policy in polynomial time with high probability.
Section~\ref{s:experiments} provides experimental comparison
between \ourmethod\ and a number of other methods on some
benchmark problems. Our results are summarized in
Section~\ref{s:discussion}. In the rest of this section, we review
the necessary preliminaries, Markov decision processes and the
exploration task.

\subsection{Markov decision processes (MDPs)}

Markov decision processes are the standard framework for RL, and
the basis of numerous extensions (like continuous MDPs, partially
observable MDPs or factored MDPs). An MDP is characterized by a
quintuple $(X, A, \mathcal R, P, \gamma)$, where $X$ is a finite
set of states; $A$ is a finite set of possible actions; $\mathcal
R: X \times A \times X \to \mathcal P_\Real$ is the reward
distribution, $R(x, a, y)$ denotes the mean value of $\mathcal
R(x,a,y)$,  $P: X \times A \times X \to [0,1]$ is the transition
function; and finally, $\gamma\in [0,1)$ is the discount rate on
future rewards. We shall assume that all rewards are nonnegative
and bounded from above by $\rmax^0$.

A (stationary) policy of the agent is a mapping \mbox{$\pi: X
\times A \to [0,1]$}. For any $x_0\in X$, the policy of the agent
and the parameters of the MDP determine a stochastic process
experienced by the agent through the instantiation $x_0, a_0, r_0,
x_1, a_1, r_1, \ldots, x_t, a_t, r_t, \ldots$

The goal is to find a policy that maximizes the expected value of
the discounted total reward. Let us define the state-action value
function (value function for short) of $\pi$ as \mbox{$Q^\pi(x,a)
:= E\Bigl( \sum_{t=0}^\infty \gamma^t r_t \Bigm| x\!=\!x_0,
a\!=\!a_0 \Bigr)$} and the optimal value function as
\[
  Q^*(x,a) := \max_{\pi} Q^\pi(x,a)
\]
for each $(x,a) \in X\times A$. Let the greedy action at $x$
w.r.t. value function $Q$ be $a_x^Q := \arg\max_a Q(x,a)$. The
greedy policy of $Q$ deterministically takes the greedy action in
each state. It is well-known that the greedy policy of $Q^*$ is an
optimal policy and $Q^*$ satisfies the Bellman equations:
\[
  Q^*(x,a) = \sum_{y} P(x,a,y) \Bigl( R(x,a,y) +
  \gamma Q^*(y,a_y^{Q^*}) \Bigr).
\]


\subsection{The exploration problem}

In the classical reinforcement learning setting, it is assumed
that the environment can be modelled as an MDP, but its parameters
(that is, $P$ and $R$) are unknown to the agent, and she has to
collect information by interacting with the environment. If too
little time is spent with the exploration of the environment, the
agent will get stuck with a suboptimal policy, without knowing
that there exists a better one. On the other hand, the agent
should not spend too much time visiting areas with low rewards
and/or accurately known parameters.

What is the optimal balance between exploring and exploiting the
acquired knowledge and how could the agent concentrate her
exploration efforts? These questions are central for RL. It is
known that the optimal exploration policy in an MDP is
non-Markovian, and can be computed only for very simple tasks like
$k$-armed bandit problems.





\section{Related literature} \label{s:literature}

Here we give a short review about some of the most important
exploration methods and their properties.

\subsection{$\epsilon$-greedy and Boltzmann exploration}

The most popular exploration method is $\epsilon$-greedy action
selection.  The method works without a model, only an
approximation of the action value function $Q(x,a)$ is needed. The
agent in state $x$ selects the greedy action $a_x^Q$ or an
explorative move with a random action with probabilities
$1-\epsilon$ and $\epsilon$, respectively. Sooner or later, all
paths with nonzero probability will have been visited many times,
so, a suitable learning algorithm can learn to choose the optimal
path. It is known, for example, that Q-learning with nonzero
exploration converges to the optimal value function with
probability 1 \citep{Littman96Generalized}, and so does SARSA
\citep{Singh00Convergence}, if the exploration rate diminishes
according to an appropriate schedule.

Boltzmann-exploration selects actions as follows: the probability
of choosing action $a$ is $\frac{\exp\bigl(Q(s,a)/T\bigr)
}{\sum_{a'\in A} \exp\bigl(Q(s,a')/T\bigr)}$, where `temperature'
$T\,(>\!\!0 )$ regulates the amount of explorative actions.
Convergence results of the $\epsilon$-greedy method carry through
to this case.

Unfortunately, for the $\epsilon$-greedy and the Boltzmann method,
exploration time may scale exponentially in the number of states
\citep{Koenig93Complexity}.

\subsection{Optimistic initial values (OIV)}

One may boost exploration with a simple trick: the initial value
of each state action  pair can be set to some overwhelmingly high
number. If a state $x$ is visited often, then its estimated value
will become more exact, and therefore, lower. Thus, the agent will
try to reach the more rarely visited areas, where the estimated
state values are still high. This method, called `exploring
starts' or `optimistic initial values', is a popular exploration
heuristic \citep{Sutton98Reinforcement}, sometimes combined with
others, e.g., the $\epsilon$-greedy exploration method. Recently,
\citet{Even-Dar01Convergence} gave theoretical justification for
the method: they proved that if the optimistic initial values are
sufficiently high, Q-learning converges to a near-optimal
solution. One apparent disadvantage of OIV is that if initial
estimations are too high, then it takes a long to fix them.

\subsection{Bayesian methods}

We may assume that the MDP (with the unknown values of $P$ and
$R$) is drawn from a parameterized distribution $\mathcal M_0$.
From the collected experience and the prior distribution $\mathcal
M_0$, we can calculate successive posterior distributions
$\mathcal M_t, t=1,2,\ldots$ by Bayes' rule. Furthermore, we can
calculate (at least in principle) the policy that minimizes the
uncertainty of the parameters \citep{Strens00Bayesian}.
\citet{Dearden00Learning} approximates the distribution of state
values directly. Exact computation of the optimal exploration
policy is infeasible and Bayesian methods are computationally
demanding even with simplifying assumptions about the
distributions, e.g., the independencies of certain parameters.

\subsection{Confidence interval estimation}

Confidence interval estimation algorithms
are between Bayesian exploration and OIV. It assumes that each
state value is drawn from an independent Gaussian distribution and
it computes the confidence interval of the state values. The agent
chooses the action with the highest upper confidence bound.
Initially, all confidence intervals are very wide, and shrink
gradually towards the true state values. Therefore, the behavior
of the technique is similar to OIV. The IEQL+ method of
\citet{Meuleau99Exploration} directly estimates confidence
intervals of $Q$-values, while \citet{Wiering98Efficient}
calculate confidence intervals for $P$ and $R$, and obtain
$Q$-value bounds indirectly. \citet{Strehl06Analysis} improve the
method and prove a polynomial-time convergence bound. Both
algorithms are called \emph{model-based interval estimation}. To
avoid confusion, we will refer to them as MBIE(WS) and MBIE(SL).

\citet{Auer06Logarithmic} give a confidence interval-based
algorithm, for which the online regret is only logarithmic in the
number of steps taken.



\subsection{Exploration Bonus Methods}

The agent can be directed towards less-known parts of the state
space by increasing the value of `interesting' states artificially
with bonuses. States can be interesting given their
\emph{frequency}, \emph{recency}, \emph{error}, etc.\
\citep{Meuleau99Exploration,Wiering98Efficient}.

The balance of exploration and exploitation is usually set by a
scaling factor $\kappa$, so that the total immediate reward of the
agent at time $t$ is $r_t + \kappa \cdot b_t(x_t,a_t,x_{t+1})$,
where $b_t$ is one of the above listed bonuses. The bonuses are
calculated by the agent and act as intrinsic motivating forces.
Exploration bonuses for a state can vary swiftly and model-based
algorithms (like prioritized sweeping or Dyna) are used for
spreading the changes effectively. Alas, the weight of exploration
$\kappa$ needs to be annealed according to a suitable schedule.

Alternatively, the agent may learn \emph{two} value functions
separately: a regular one, $Q^r_t$ which is based on the rewards
$r_t$ received from the environment, and an exploration value
function $Q_t^e$ which is based on the exploration bonuses. The
agent's policy will be greedy with respect to their combination
$Q^r_t + \kappa Q_t^e$. Then the exploration mechanism may remain
the same, but several advantages appear. First of all, the changes
in $\kappa$ take effect \emph{immediately}. As an example, we can
immediately switch off exploration by setting $\kappa$ to 0.
Furthermore, $Q^r_t$ may converge even if $Q_t^e$ does not.

Confidence interval estimation can be phrased as an exploration
bonus method: see IEQL+ \citep{Meuleau99Exploration} or MBIE-EB
\citep{Strehl06Analysis}. \citet{Even-Dar01Convergence} have shown
that $\epsilon$-greedy and Boltzmann explorations can be
formulated as exploration bonus methods although rewards are not
propagated through the Bellman equations.

\subsection{$E^3$ and R-max}

The \emph{Explicit explore or exploit} ($E^3$) algorithm of
\citet{Kearns98Near-Optimal} and its successor, R-max
\citep{Brafman01R-MAX} were the first algorithms that have
polynomial time bounds for finding near-optimal policies. R-max
collects statistics about transitions and rewards. When visits to
a state enable high precision estimations of real transition
probabilities and rewards then state is declared \emph{known}.
R-max also maintains an approximate model of the environment.
Initially, the model assumes that all actions in all states lead
to a (hypothetical) maximum-reward absorbing state. The model is
updated each time when a state becomes known. The optimal policy
of the model is either the near-optimal policy in the real
environment or enters a not-yet-known state and collects new
information.

\section{Construction of the algorithm} \label{s:algorithm}

Our agent starts with a simple, but overly optimistic model. By
collecting new experiences, she updates her model, which becomes
more realistic. The value function is computed over the
approximate model with (asynchronous) dynamic programming. The
agent always chooses her action greedily w.r.t. her value
function. Exploration is induced by the optimism of the model:
unknown areas are believed to yield large rewards. Algorithmic
components are detailed below.



\textbf{Separate exploration values.} Similarly to the approach of
\citet{Meuleau99Exploration}, we shall separate the `true' state
values from exploration values. Formally, the value function has
the form
\[
  Q(x,a) = Q^r(x,a) + Q^e(x,a)
\]
for all $(x,a)\in X \times A$, where $Q^r$ and $Q^e$ will
summarize external and exploration rewards, respectively.

\textbf{`Garden of Eden' state.} Similarly to R-max, we introduce
a new hypothetical `garden of Eden' state $x_E$, and assume an
extended state space $X'=X \cup \{x_E\}$. Once there, then,
according to the inherited model, the agent remains in $x_E$
indefinitely and receives $\ro$ reward for every step, which may
exceed \mbox{$\rmax^0=:\max_{x,a,y} R(x,a,y)$}, the maximal reward
of the original environment.

\textbf{Model approximation.} The agent builds an approximate
model of the environment. For each $x,y\in X$ and $a\in A$, let
$N_t(x,a)$, $N_t(x,a,y)$, and $C_t(x,a,y)$ denote the number of
times when $a$ was selected in $x$ up to step $t$, the number of
times when transition $x\stackrel{a}{\rightarrow} y$ was
experienced, and the sum of external rewards for
$x\stackrel{a}{\rightarrow} y$ transitions, respectively. With
these notations, the approximate model parameters are
\[
  \hat{P}_t(x,a,y) = \frac{N_t(x,a,y)}{N_t(x,a)} \textrm{ and }  \hat{R}_t(x,a,y) =
  \frac{C_t(x,a,y)}{N_t(x,a,y)}.
\]
Suitable initializations of $N_t(x,a)$, $N_t(x,a,y)$ and
$C_t(x,a,y)$ will ensure that the ratios are well-defined
everywhere. The exploration rewards are defined as
\[
  R^e(x,a,y) := \left\{%
\begin{array}{ll}
    \ro, & \hbox{if $y=x_E$;} \\
    0, & \hbox{if $y\neq x_E$,} \\
\end{array}%
\right.
\]
for each $x,y\in X\cup\{x_E\}$, $a\in A$, and are not modified
during the course of learning.

\textbf{Optimistic initial model.} The initial model assumes that
$x_E$ has been reached once for each state-action pairs: for each
$x\in X \cup \{x_E\}$, $y\in X$ and $a\in A$,
\[
  \begin{array}{ll}
    N_0(x,a) = 1, \\
     N_0(x,a,y) = 0, & C_0(x,a,y) = 0. \\
     N_0(x,a,x_E) = 1, & C_0(x,a,x_E) = 0. \\
  \end{array}
\]
Then, the optimal initial value function equals
\[
  Q_0(x,a) = Q_0^r (x,a) + Q_0^e (x,a) = 0 + \frac{1}{1-\gamma}\ro := \vo
\]
for each $(x,a)\in X'\times A$, analogously to OIV.

\textbf{Dynamic programming.}  Both value functions can be updated
using the approximate model. For each $x\in X$, let $a_x$ be the
greedy action according to the combined value function, i.e.,
\[
  a_x := \arg\max_{a\in A} \bigl( Q^r(x,a) + Q^e(x,a) \bigr).
\]
The dynamic programming equations for the value function
components are
\begin{eqnarray*}
  Q^r_{t+1}(x,a) \hspace{-3mm} &:=& \hspace{-3mm}\sum_{y\in X} \hat{P}_t(x,a,y) \left( \hat{R}_t(x,a,y) + \gamma Q^r_t(y,a_y)
  \right)\\
  Q^e_{t+1}(x,a)\hspace{-3mm} &:=& \hspace{-3mm} \gamma \sum_{y\in X} \hat{P}_t(x,a,y)
  Q^e_t(y,a_y) \\
  & & + \,\,  \hat{P}_t(x,a,x_E)  \vmax .
\end{eqnarray*}
Episodic tasks can be handled as usual way; we introduce an
absorbing final state with 0 external reward.

\textbf{Asynchronous update.} The algorithm can be online, if
instead of full update sweeps over the state space updates are
limited to state set $L_t$ in the `neighborhood' of the agent's
current state. Neighborhood is restricted by computation time
constraints; any asynchronous dynamic programming algorithm
suffices. It is implicitly assumed that the current state is
always updated, i.e., $x_t\in L_t$. In this paper, we used the
improved prioritized sweeping algorithm of
\citet{Wiering98Efficient}.

\textbf{Putting it all together.} The method is summarized as
Algorithm 1.

\begin{algorithm*}[t]
   \caption{The Optimistic initial model algorithm}
   \label{alg:example}
\begin{algorithmic}
   \STATE {\bfseries Input:} $x_0 \in X$ initial state, $\epsilon>0$ required
   precision, optimism parameter $\rmax$
   \STATE {\bfseries Model initialization:} $t:=0$; $\forall x,y\in X, \forall a\in A$:
   \STATE $N(x,a,y) := 0$, $N(x,a,x_E) := 1$, $N(x,a) := 1$, $C(x,a,y) := 0$, $Q^r(x,a) := 0$, $Q^e(x,a) :=
   \rmax/(1-\gamma)$;
   \REPEAT
   \STATE $a_t:=$ greedy action w.r.t. $Q^r+Q^e$; apply $a_t$ and observe $r_t$ and $x_{t+1}$
   \STATE $C(x_t,a_t,x_{t+1}) := C(x_t,a_t,x_{t+1})+r_t
   $;  $N(x_t,a_t,x_{t+1}) := N(x_t,a_t,x_{t+1})+1 $; $N(x_t,a_t) := N(x_t,a_t)+1 $
   \STATE $L_t$ := list of states to be updated
   \FOR{{\bfseries each} $x\in L_t$ }
   \STATE $Q^r_{t+1}(x,a) := \sum_{y\in X} \hat{P}(x,a,y) \left( \hat{R}(x,a,y) + \gamma Q^r_t(y,a_y)
  \right)$
   \STATE $Q^e_{t+1}(x,a) :=  \hat{P}(x,a,x_E)  \rmax/(1-\gamma)  + \gamma \sum_{y\in X} \hat{P}(x,a,y)
  Q^e_t(y,a_y).$
   \ENDFOR
   \STATE $t := t+1$
   \UNTIL{Bellman-error$>\epsilon$}
\end{algorithmic}
\end{algorithm*}

\section{Analysis} \label{s:analysis}

In the first part of this section, we analyze the similarities and
differences between various exploration methods, with an emphasis
on \ourmethod. Based on this analysis, we sketch the proof that
\ourmethod\ finds a near-optimal policy in polynomial time.

\subsection{Relationship to other methods}

`Optimism in the face of uncertainty' is a common point in
exploration methods: the agent believes that she can obtain extra
rewards by reaching the unexplored parts of the state space.

Note that as far as the combined value function $Q$ is concerned,
\ourmethod\ is an asynchronous dynamic programming method
augmented with model approximation.

\textbf{Optimistic initial values.} Apparently, \ourmethod\ is the
model-based extension of the OIV heuristic. Note however, that
optimistic initialization of $Q$-values is not effective with a
model: the more updates are made, the less effect the
initialization has and it fully diminishes if value iteration is
run until convergence. Therefore, naive combination of OIV and
model construction is contradictory: the number of DP-updates
should be kept low in order to save the initial boost, but it
should be as high as possible in order to propagate the real
rewards quickly.

\ourmethod\ resolves this paradox by moving the optimism into the
model. The optimal value function of the initial model is $Q_0
\equiv \vmax$, corresponding to OIV. However, DP updates can not,
but only model updates may lower the exploration boost.

Note that we can set the initial model value as high as we like,
but we do not have to wait until the initial boost diminishes,
because $Q^r$ and $Q^e$ are separated.

\textbf{R-max.} The `Garden of Eden' state $x_E$ of \ourmethod\ is
identical to the fictitious max-reward absorbing state of
\textsc{R-max} (and $E^3$). In both cases, the agent's model tells
that all unexplored $(x,a)$ pairs lead to $x_E$.
\mbox{\textsc{R-max}}, however, updates the model only when the
transition probabilities and rewards are known with high
precision, which is only after many visits to $(x,a)$. In
contrast, \ourmethod\ updates the model after each single visit,
employing each bit of experience as soon as it is obtained. As a
result, the approximate model can be used long before it becomes
accurate.

\textbf{Exploration bonus methods.} The extra reward offered by
the Garden of Eden state can be understood as an exploration
bonus: for each visit of the pair $(x,a)$, the agent gets the
bonus $b_t (x,a) = \frac{1}{N_t(x,a)}\bigl( \vmax - Q_t(x,a)
\bigr)$. It is insightful to contrast this formula with those of
the other methods like the \emph{frequency-based} bonus $b_t= -
\alpha \cdot N_t(x,a)$ or the \emph{error-based} bonus $b_t=
\alpha \cdot \bigl| Q_{t+1}(x,a) - Q_{t}(x,a) \bigr|$.

\textbf{Model-based interval exploration.} The exploration bonus
form of the MBIE method of \citet{Strehl05Theoretical} sets $b_t =
\frac{\alpha}{N_t(x,a)}$. MBIE-EB is not an ad-hoc method: the
form of the bonus comes from confidence interval estimations. The
comparison to MBIE-EB will be especially valuable, as it converges
in polynomial-time and the proof can be transported to \ourmethod\
with slight modifications.

\subsection{Polynomial-time convergence}

\begin{thm}
For any $\epsilon>0, \,
\delta>0$, $\epsilon_1 := \epsilon/6$, $\epsilon_2 := \frac{(1-\gamma)^2}{\xsize \left( 1-\gamma +
  \rmax^0
\right)}\cdot \epsilon_1$, $H:= \frac{1}{1-\gamma} \ln \frac{\rmax^0}{\epsilon_1
(1-\gamma)}$, $m := \frac{2\max\{1,\rmax^0\}^2}{\epsilon_2^2} \ln
\frac{8}{\delta}$, OIM converges almost surely to a near-optimal
policy in polynomial time if started with $\rmax =
\frac{2(\rmax^0)^2\ln(2\left|X\right|
        \left|A\right| m /\delta)}{\epsilon_1
        (1-\gamma)^3}$,
that is, with probability $1-\delta$, the number of timesteps
where $Q^{\pi^\ourmethod}(x_t,a_t) > Q^*(x_t,a_t)-\epsilon$ does
not hold, is at most $\frac{2m \left|X\right| \left|A\right| H
\rmax^0}{\epsilon_1(1-\gamma)} \ln \frac{4}{\delta} $.
\end{thm}

The proof can be found in the Appendix.

\section{Experiments} \label{s:experiments}

To assess the practical utility of \ourmethod, we compared its
performance to other exploration methods. Experiments were run on
several small benchmark tasks challenging exploration algorithms.

For fair comparisons, benchmark problems were taken from the
literature without changes, nor did we change the experimental
settings or the presentation of experimental data. It also means
that the presentation format varies for different benchmarks.

\subsection{RiverSwim and SixArms}

The first two benchmark problems, \emph{RiverSwim} and
\emph{SixArms}, were taken from \citet{Strehl06Analysis}.

The \emph{RiverSwim} MDP has 6 states, representing the position
of the agent in a river. The agent has two possible actions: she
can swim either upstream or downstream. Swimming down is always
successful, but swimming up succeeds only with a 30\% chance and
there is a 10\% chance of slipping down. The lowermost position
yields $+5$ reward per step, while the uppermost position yields
$+10000$.


The \emph{SixArms} MDP consists of a central state and six `payoff
states'. In the central state, the agent can play 6 one-armed
bandits. If she pulls arm $k$ and wins, she is transferred to
payoff state $k$. Here, she can get a reward in each step, if she
chooses the appropriate action. The winning probabilities range
from 1 to 0.01, while the rewards range from 50 to 6000 (for the
exact values, see \npcite{Strehl06Analysis}).


Data for $E^3$, {\sc R-max}, {\sc MBIE} and {\sc MBIE-EB} are
taken from \citet{Strehl06Analysis}. Parameters of all four
algorithms were chosen optimally. Following a coarse search in
parameter space, the $\ro$ parameter for \ourmethod\ was set to
$2000$ for \emph{RiverSwim} and to $10000$ for \emph{SixArms}.
State spaces are small and value iteration instead of prioritized
sweeping was completed in each step.

On both problems, each algorithm ran for 5000 time steps and the
undiscounted total reward was recorded. The averages and 95\%
confidence intervals are calculated over 1000 test runs
(Tables~\ref{t:riverswim} and \ref{t:sixarms}).

\begin{table}
\caption{Results on the \emph{RiverSwim} task. }\label{t:riverswim}
\begin{center}
{
\begin{tabular}{l || @{\ \ }r @{$\cdot 10^6 \ \pm$}r @{$\;\cdot 10^6$} }
{Method}          & \multicolumn{2}{c}{Cumulative reward}\\[2pt]
\hline \rule{0pt}{15pt}%
{$E^3$}          & 3.020 & 0.027 \\
{\sc R-max}      & 3.014 & 0.039 \\
{\sc MBIE(SL)}       & 3.168 & 0.023 \\
{\sc MBIE-EB}    & 3.093 & 0.023 \\
{\ourmethod}     & 3.201 & 0.016 \\
\hline
\end{tabular}
}
\end{center}
\vskip -0.1in
\end{table}

\begin{table}
\caption{Results on the \emph{SixArms} task.  } \label{t:sixarms}
\begin{center}
{
\begin{tabular}{l || @{\ \ }r @{$\cdot 10^6 \ \pm$ }r @{$\;\cdot 10^6$} }
{Method}          & \multicolumn{2}{c}{Cumulative reward}\\[2pt]
\hline \rule{0pt}{15pt}%
{$E^3$}          & 1.623 & 0.244 \\
{\sc R-max}      & 2.819 & 0.256 \\
{\sc MBIE(SL)}       & 9.205 & 0.559 \\
{\sc MBIE-EB}    & 9.486 & 0.587 \\
{\ourmethod}     & 10.007 & 0.654 \\
%
\hline
\end{tabular}
}
\end{center}
\vskip -0.1in
\end{table}


\subsection{$50\times50$ maze with subgoals}

Another benchmark problem, \emph{MazeWithSubgoals}, was suggested
by \citet{Wiering98Efficient}. The agent has to navigate in a
$50\times 50$ maze from the start position at $(2,2)$ to the goal
(with +1000 reward) at the opposite corner $(49,49)$. There are
suboptimal goals (with +500 reward) at the other two corners. The
maze has blocked places and punishing states ($-10$ reward), set
randomly in 20-20\% of the squares. The agent can move in four
directions, but with a 10\% chance, its action is replaced by a
random one. If the agent tries to move to a blocked state, it gets
a reward of $-2$. Reaching any of the goals resets the agent to
the start state. In all other cases, the agent gets a $-1$ reward
for each step.


Each algorithm was run on 20 different mazes for 100,000 steps.
After every 1000 steps, we tested the learned value functions by
averaging 20 test runs, in each one following the greedy policy
for 10,000 steps, and averaging cumulated (undiscounted) rewards.
We measured the number of test runs needed for the algorithms to
learn to collect 95\%, 99\% and 99.8\% of the maximum possible
rewards in 100,000 steps, and the number of steps this takes on
average, if the algorithms can meet the challenge.

The algorithms that we compared were the recency based and frequency based
exploration bonus methods, two versions of $\epsilon$-greedy exploration,
MBIE(WS) and \ourmethod. All exploration rules applied the improved prioritized
sweeping of \citet{Wiering98Efficient}. \ourmethod's $\ro$ was set to 1000. The
results are summarized in Table~\ref{t:wmaze2}.

\begin{table}
\caption{Results on the \emph{MazeWithSubgoals} task. The number
of steps required to learn $p$-optimal policies ($p$=0.95, 0.99,
0.998) on the $50\times50$ maze task with suboptimal goals. In
parentheses: how many runs out of 20 have found the goal. `$k$'
stands for 1000.} \label{t:wmaze2}
\begin{center}
{
\begin{tabular}{l || @{\ \ }r|@{\ \ }r|@{\ \ }r c}
{Method}          & 95\% & 99\% & 99.8\%\\[2pt]
\hline \rule{0pt}{15pt}%
{\sc $\epsilon$-greedy, $\epsilon=0.2$}          & -- \ (0)          & -- \ (0)      & -- \ (0) \\
{\sc $\epsilon$-greedy, $\epsilon=0.4$}          & 43{k} \, (4)       & 52{k} \, (4)   & 68{k} \, (4) \\
{\sc Recency-bonus}             & 27{k} (19)        & 55{k} (18)    & 69{k} \, (9)\\
{\sc Freq.-bonus}           & 24{k} (20)         & 50{k} (16)     & 66{k} (10) \\
{\sc MBIE(WS)}                      & 25{k} (20)         & 42{k} (19)     & 66{k} (18) \\
{\ourmethod}                     & 19{k} (20)         & 29{k} (20)     & 31{k} (20) \\
\hline
\end{tabular}
}
\end{center}
\vskip -0.1in
\end{table}

\subsection{Chain, Loop and FlagMaze}

The next three benchmark MDPs, the \emph{Chain}, \emph{Loop} and
\emph{FlagMaze} tasks were investigated, e.g., by
\citet{Meuleau99Exploration}, \citet{Strens00Bayesian} and
\citet{Dearden00Learning}. In the \emph{Chain} task, 5 states are
lined up along a chain. The agent gets +2 reward for being in
state 1 and +10 for being in state 5. One of the actions advances
one state ahead, the other one resets the agent to state 1. The
\emph{Loop} task has 9 states in two loops (arranged in a
8-shape). Completing the first loop (using any combination of the
two actions) yields +1 reward, while the second loop yields +2,
but one of the actions resets the agent to the start. The
\emph{FlagMaze} task consists of a $6\times 7$ maze with several
walls, a start state, a goal state and 3 flags. Whenever the agent
reaches the goal, her reward is the number of flags collected.

The following algorithms were compared: Q-learning with
variance-based and TD error-based exploration bonus (model-free
variants), $\epsilon$-greedy exploration, Boltzmann exploration,
IEQL+, Bayesian Q-learning, Bayesian DP and \ourmethod. Data were
taken from \citet{Meuleau99Exploration}, \citet{Strens00Bayesian}
and \citet{Dearden00Learning}. According to the sources,
parameters for all algorithms were set optimally. \ourmethod's
$\ro$ parameter was set to 0.5, 10 and 0.005 for the three tasks,
respectively.

Each algorithm ran for 8 learning phases. The total cumulated
reward over each learning phase was measured. One phase lasted for
1000 steps for the first two tasks and 20,000 steps for the
\emph{FlagMaze} task. We carried out 256 parallel runs for the
first 2 tasks and 20 for the third one.

\begin{table}
\caption{Average accumulated rewards on the \emph{Chain} task.
Optimal policy gathers 3677.} \label{t:chain}
\begin{center}
{
\begin{sc}
\begin{tabular}{l||c|c|c}
Method & Phase 1 & Phase 2 & Phase 8 \\[2pt]
\hline \rule{0pt}{12pt}%

QL+var.-bonus      & --             & 2570\footnotemark[1]    & --       \\
QL+err.-bonus      & --             & 2530\footnotemark[1]    & --       \\
QL $\epsilon$-greedy     & 1519      & 1611      & 1602 \\
QL Boltzmann        & 1606     & 1623     & --      \\
IEQL+               & 2344     & 2557     & --      \\
Bayesian QL        & 1697    & 2417    & --      \\
Bayesian DP\footnotemark[2]         & 3158     & 3611     & 3643 \\

\ourmethod          & 3510    & 3628    & 3643 \\
\end{tabular}
\end{sc}
}
\end{center}
\end{table}


\begin{table}
\caption{Average accumulated rewards on the \emph{Loop} task.
Optimal policy gathers 400.} \label{t:loop}
\begin{center}
{
\begin{sc}
\begin{tabular}{l||c|c|c}
Method & Phase 1 & Phase 2 & Phase 8 \\[2pt]
\hline \rule{0pt}{12pt}%

QL+var.-bonus       & --             & 179\footnotemark[1]    & --       \\ 
QL+err.-bonus       & --             & 179\footnotemark[1]    & --       \\
QL $\epsilon$-greedy     & 337       & 392       & 399 \\
QL Boltzmann        & 186       & 200       & -- \\
IEQL+               & 264       & 293       & --      \\
Bayesian QL       & 326      & 340      &  --      \\
Bayesian DP\footnotemark[2]         & 377       & 397   & 399 \\

\ourmethod          & 393      & 400    & 400 \\
\end{tabular}
\end{sc}
}
\end{center}
\end{table}

\begin{table}
\caption{Average accumulated rewards on the \emph{FlagMaze} task.
Optimal policy gathers approximately 1890.} \label{t:maze}
\begin{center}
{
\begin{sc}
\begin{tabular}{l||c|c|c}
Method & Phase 1 & Phase 2 & Phase 8 \\[2pt]
\hline \rule{0pt}{12pt}%

QL $\epsilon$-greedy      & 655        & 1135       & 1147 \\
QL Boltzmann        & 195        & 1024       & -- \\
IEQL+               & 269         & 253         & -- \\
Bayesian QL       & 818        & 1100       & --  \\
Bayesian DP\footnotemark[2]         & 750         & 1763        & 1864 \\

\ourmethod          & 1133        & 1169       & 1171 \\
\end{tabular}
\end{sc}
}
\end{center}
\end{table}
 \footnotetext[1]{\normalsize Results for Phase 5.}
 \footnotetext[2]{\normalsize Augmented with limited amount of pre-wired knowledge (the list of successor
states).}

\section{Summary of the results} \label{s:discussion}

We proposed a new algorithm for exploration and reinforcement
learning in Markov decision processes. The algorithm integrates
concepts from other advanced exploration methods. The key
component of our algorithm is an optimistic initial model. The
optimal policy according to the agent's model will either explore
new information that helps to make the model more accurate, or
follows a near-optimal path. The extent of optimism regulates the
amount of exploration. We have shown that with a suitably
optimistic initialization, our algorithm finds a near-optimal
policy in polynomial time. Experiments were conducted on a number
of benchmark MDPs. According to the experimental results our novel
method is robust and compares favorably to other methods.

%
%

%

\subsection*{Acknowledgments}

We are grateful to Marcus Hutter for calling our attention to an error regarding the
$\rmax$ bound, and to one of the reviewers for his helpful comments.
This research has been supported by the EC FET `New Ties' Grant
FP6-502386 and NEST `PERCEPT' Grant FP6-043261. Opinions and
errors in this manuscript are the author's responsibility, they do
not necessarily reflect those of the EC or other project members.

\appendix

\section{Proof of Polynomial-time convergence}


For the proof, we shall follow the technique of
\citet{Kearns02Near-Optimal} and \citet{Strehl06Analysis}, and
will use the shorthands [KS] and [SL] for referring to them. We
will proceed by a series of lemmas.

Throughout the proof, note the difference between $\rmax$ and $\rmax^0$. Value estimates of our
model start from $\rmax$. However, all actual rewards observed by
the agent are bounded by $\rmax^0$, which is smaller than $\rmax$.

\begin{lem} (Azuma's Lemma) If the random variables $X_1, X_2, \ldots$ form a
martingale difference sequence, meaning that $E[X_k|X_1,X_2,
\ldots,X_{k-1}] = 0$ for all $k$, and $|X_k| \leq b$ for each $k$,
then
\[
  \Pr \left[ \sum_{i=1}^k X_i \geq a \right] \leq \exp \left( -
  \frac{a^2}{2b^2k}\right)
\]
and
\[
  \Pr \left[ \left|\sum_{i=1}^k X_i\right| \geq a \right] \leq 2\exp \left( -
  \frac{a^2}{2b^2k}\right)
\]
\end{lem}

The following lemma is similar to Lemma 5 of {[KS]} (with the
modification that $R(x,a,y)$ values are learnt instead of
$R(x,a)$-values, and tells that if a state-action pair is visited
many times, then its parameter estimates become accurate.

\begin{lem} \label{lem:knowingPR}
Consider an MDP $M =
(X,A,P,R,\gamma)$, and let $(x,a)$ be a state-action pair that has
been visited at least $m$ times. Let $\hat{P}(x,a,y)$ and
$\hat{R}(x,a,y)$ denote the obtained empirical estimates, let
$\epsilon>0$ and $\delta>0$ be arbitrary positive values. If
\[
 m \geq \frac{2\max\{1,\rmax^0\}^2}{\epsilon^2} \ln \frac{2}{\delta} ,
\]
then for all $y\in X$,
\begin{eqnarray*}
  \left| P(x,a,y) R(x,a,y) - \hat P(x,a,y)\hat R(x,a,y) \right| &\leq& \epsilon \qquad\textrm{and} \\
  \left| P(x,a,y) - \hat P(x,a,y) \right| &\leq& \epsilon
\end{eqnarray*}
holds with probability at least $1-\delta$.
\end{lem}
 \bproof
Suppose that $(x,a)$ is visited $k$ times at steps $t_1, \ldots,
t_k$. Define the random variables
\[
  Z_i(y) = \left\{%
\begin{array}{ll}
    1, & \hbox{if $x_{t_i+1} = y$;} \\
    0, & \hbox{otherwise.} \\
\end{array}%
\right.
\]
Clearly, $E[Z_i(y)] = P(x,a,y)$ and $Z_i(y)-P(x,a,y)$ is a
martingale, so we can apply Azuma's lemma with $a = k\epsilon$ to
get
\[
  \Pr \left[ \left| \frac{1}{k}\sum_{i=1}^k Z_i - P(x,a,y) \right| \geq \epsilon \right] \leq 2\exp \left( -
  \frac{\epsilon^2 k}{2}\right)
  \leq 2\exp \left( -\frac{\epsilon^2 m}{2}\right).
\]
The right-hand side is less than $\delta$ for
\[
m \geq \frac{2}{\epsilon^2} \ln \frac{2}{\delta}.
\]
Similarly, define the random variables
\[
  W_i(y) = \left\{%
\begin{array}{ll}
    r_{t_i+1}, & \hbox{if $x_{t_i+1} = y$;} \\
    0, & \hbox{otherwise.} \\
\end{array}%
\right.
\]
In this case, $E[W_i(y)] = P(x,a,y) R(x,a,y)$,
$W_i(y)-P(x,a,y)R(x,a,y)$ is a martingale and is bounded by
$\rmax^0$ (note that we are considering only states $x,y\in X$, that is,
the garden-of-Eden state $x_E$ is excluded. Therefore, $\rmax^0$ is indeed
an upper bound on $R(x,a,y)$), so we can apply Azuma's lemma with $a = k\epsilon$ to
get
\[
  \Pr \left[ \left| \frac{1}{k}\sum_{i=1}^k W_i - P(x,a,y)R(x,a,y) \right| \geq \epsilon \right] \leq 2\exp \left( -
  \frac{\epsilon^2 k}{2(\rmax^0)^2}\right)
  \leq 2\exp \left( -\frac{\epsilon^2 m}{2(\rmax^0)^2}\right).
\]
The right-hand side is less than $\delta$ for
\[
m \geq \frac{2(\rmax^0)^2}{\epsilon^2} \ln \frac{2}{\delta}.
\]
Unifying the two requirements for $m$ completes the proof of the
lemma.
 \eproof

The following is a minor modification of {[KS]} lemma 4, and [SL]
Lemma 1. The result tells that if the parameters of two MDPs are
very close to each other, then the value functions in the two MDPs
will also be similar.

\begin{lem} \label{lem:simulation}
Let $\epsilon>0$, and consider two MDPs $M = (X,A,P,R,\gamma)$ and
$\bar M = (X,A,\bar P,\bar R,\gamma)$ that differ only in their
transition and reward functions, furthermore, their difference is
bounded:
\begin{eqnarray*}
  \left| P(x,a,y)R(x,a,y) - \bar P(x,a,y)\bar R(x,a,y) \right| &\leq& \epsilon' \qquad\textrm{and} \\
  \left| P(x,a,y) - \bar P(x,a,y) \right| &\leq& \epsilon'
\end{eqnarray*}
for all $(x,a,y)\in X\times A\times X$ and
\begin{eqnarray*}
  \epsilon' &:=& \frac{(1-\gamma)^2}{\xsize \left( 1-\gamma + \rmax^0
\right)}\cdot \epsilon.
\end{eqnarray*}
Then for any policy $\pi$ and any $(x,a)\in X\times A$,
\[
  \left| Q^\pi(x,a) - \bar Q^\pi(x,a) \right| \leq \epsilon.
\]
\end{lem}
 \bproof
Let $\Delta := \max_{(x,a)\in X\times A} |Q^\pi(x,a) - \bar
Q^\pi(x,a)|$, and note that for any $x\in X$,
\begin{eqnarray*}
 |V^\pi(x) - \bar V^\pi(x)| &=& \left| \sum_a \pi(x,a) (Q^\pi(x,a) - \bar
Q^\pi(x,a)) \right| \leq \sum_a \pi(x,a) \Delta = \Delta
\end{eqnarray*}
For a fixed $(x,a)$ pair,
\begin{eqnarray*}
 && \Delta = \left| Q^\pi (x,a) - \bar Q^\pi(x,a) \right| \\
 && = \left| \sum_{y\in X} P(x,a,y) \Bigl(R(x,a,y) + \gamma V^\pi(y) \Bigr)
    - \sum_{y\in X} \bar P(x,a,y) \Bigl(\bar R(x,a,y) + \gamma \bar V^\pi(y) \Bigr) \right| \\
 && \leq \sum_{y\in X}  \left|P(x,a,y)R(x,a,y)-\bar P(x,a,y) \bar R(x,a,y)  \right| \\
 && \quad + \left| \sum_{y\in X} \Bigl[P(x,a,y)-\bar P(x,a,y)\Bigr] \Bigl(\gamma V^\pi(y) \Bigr) \right|
     + \left| \sum_{y\in X} \bar P(x,a,y) \Bigl( \gamma \Bigl[V^\pi(y) - \bar V^\pi(y)\Bigr] \Bigr) \right| \\
 && \leq \xsize \epsilon' + \sum_{y\in X} \epsilon' \Bigl| \gamma V^\pi(y) \Bigr|
    +  \sum_{y\in X} \bar P(x,a,y) (\gamma \Delta ) \\
 && \leq  \xsize \epsilon' + \xsize\epsilon' \frac{\rmax^0}{1-\gamma}
    +    \gamma \Delta.
\end{eqnarray*}
Therefore,
\[
\Delta \leq \frac{\xsize \epsilon' \left( 1-\gamma + \rmax^0
\right)}{(1-\gamma)^2} = \epsilon
\]
 \eproof

Let us introduce a modified version of \ourmethod\ that behaves
exactly like the old one, except that in each $(x,a)$ pairs, it
performs at most $m$ updates. If a pair is visited more than $m$
times, the modified algorithm leaves the counters unchanged.

The following result is a modification of {[SL]}'s Lemma 7.

\begin{lem} \label{lem:betaineq}
Suppose that the modified \ourmethod\ (stopping after $m$ updates)
is executed on an MDP $M= (X,A,P,R,\gamma)$ with
\begin{eqnarray*}
 m &:=&
\frac{2\max\{1,\rmax^0\}^2}{\epsilon^2} \ln \frac{2}{\delta} , \\
  \beta &:=& \frac{\rmax^0}{1-\gamma} \sqrt{2\ln(2\left|X\right|
\left|A\right| m /\delta)}.
\end{eqnarray*}
Then, with probability at least $1 - \delta$,
\[
 Q^*(x,a) -
\sum_{y\in X} \hat{P}_t(x,a,y) \left[ \hat{R}_t (x,a,y) + \gamma
V^*(y) \right]
 \leq \beta/\sqrt{k}
\]
for all $t=1,2,\ldots$
\end{lem}
\bproof Fix a state-action pair $(x,a)$ and suppose that it has
been visited $k\leq m$ times until time step $t$, at steps $t_1,
\ldots, t_k$. Define the random variables $X_1, \ldots, X_k$ by
\[
  X_i := r_{t_i} + \gamma V^*(x_{t_i+1}).
\]
Note that $E[X_i] = Q^*(x,a)$ and $0\leq X_i \leq
\rmax^0/(1-\gamma)$ for all $i = 1, \ldots, k$, and the sequence
$Q^*(x,a) - X_i$ is a martingale difference sequence. Applying
Azuma's lemma yields
\begin{eqnarray}
  \Pr \left[ E[X_1] - \frac{1}{k}\sum_{i=1}^k X_i \geq a/k \right] &\leq&
  \exp \left( - \frac{a^2(1-\gamma)^2}{2(\rmax^0)^2k}\right)
  \label{e:azuma1}
\end{eqnarray}
for any $a$. Let the right-hand side be equal to
$\frac{\delta}{2\left|X\right| \left|A\right| m}$, corresponding
to
\[
  a = \beta\sqrt{k}
\]
with
\[
  \beta := \rmax^0/(1-\gamma) \sqrt{2\ln(2\left|X\right|
  \left|A\right| m /\delta)}.
\]

Note that by the construction of the \ourmethod\ algorithm,
\begin{eqnarray*}
  &&\sum_{y\in X'} \hat{P}_t(x,a,y) \left[ \hat{R}_t (x,a,y) + \gamma V^*(y) \right]
   \\
  && = \sum_{y\in X: N_t(x,a,y)>0} \frac{N_t(x,a,y)}{N_t(x,a)} \left[ \frac{C_t(x,a,y)}{N_t(x,a,y)} + \gamma V^*(y) \right]
   + \frac{1}{N_t(x,a)} [ \rmax + \gamma V^*(x_{GOE})]   \\
  && =  \frac{N_t(x,a)-1}{N_t(x,a)}  \sum_{y\in X: N_t(x,a,y)>0} \frac{N_t(x,a,y)}{N_t(x,a)-1} \left[ \frac{C_t(x,a,y)}{N_t(x,a,y)} + \gamma V^*(y) \right]
   + \frac{1}{N_t(x,a)} \frac{\rmax}{1-\gamma}    \\
  && =  \frac{k}{k+1}  \sum_{y\in X: N_t(x,a,y)>0} \frac{N_t(x,a,y)}{k} \left[ \frac{C_t(x,a,y)}{N_t(x,a,y)} + \gamma V^*(y) \right]
   + \frac{1}{k+1} \frac{\rmax}{1-\gamma}    \\
  && =  \frac{k}{k+1} \cdot  \frac{1}{k} \sum_{i=1}^k  X_i
   + \frac{1}{k+1} \frac{\rmax}{1-\gamma},
\end{eqnarray*}
where we exploited the fact that $k = N_t(x,a)-1$. Therefore,
\begin{eqnarray*}
  \frac{1}{k} \sum_{i=1}^k X_i = \frac{k+1}{k}  \sum_{y\in X'} \hat{P}_t(x,a,y) \left[ \hat{R}_t (x,a,y) + \gamma V^*(y) \right]
  - \frac{1}{k} \cdot \frac{\rmax}{1-\gamma}.
\end{eqnarray*}
Substituting this to (\ref{e:azuma1}), we get that
\begin{eqnarray*}
 Q^*(x,a) - \frac{k+1}{k}  \sum_{y\in X'} \hat{P}_t(x,a,y) \left[ \hat{R}_t (x,a,y) + \gamma V^*(y) \right]
  + \frac{1}{k} \cdot \frac{\rmax}{1-\gamma} < \beta/\sqrt{k}
\end{eqnarray*}
with high probability, but we will use only the slightly looser
inequality
\begin{eqnarray}
 Q^*(x,a) -   \sum_{y\in X} \hat{P}_t(x,a,y) \left[ \hat{R}_t (x,a,y) + \gamma V^*(y) \right]
 \leq \beta/\sqrt{k}.
  \label{e:betaineq1}
\end{eqnarray}

For each $(x,a)$, the modified \ourmethod\ algorithm changes the
parameters at most $m$ times, which is at most $m \left|X\right|
\left|A\right|$ changes in total. Each different approximation
fails with probability less than $\frac{\delta}{2\left|X\right|
\left|A\right| m}$, so, by the union bound, the total probability
that (\ref{e:betaineq1}) fails (at any time, for any state-action
pair) is still less than $\delta/2$.

\eproof

The following result shows that the modified \ourmethod\ algorithm
preserves the optimism of the value function with high
probability.

\begin{lem} \label{lem:optimismpreserved}
Let $\epsilon_1>0$ and suppose that the modified \ourmethod\ is
executed on an MDP $M=(X,A,P,R,\gamma)$ with
\[
  \rmax \geq \frac{\beta^2}{\epsilon_1}
\]
where
\begin{eqnarray*}
 m &:=&
\frac{2\max\{1,\rmax^0\}^2}{\epsilon^2} \ln \frac{2}{\delta} , \\
  \beta &:=& \frac{\rmax^0}{1-\gamma} \sqrt{2\ln(2\left|X\right|
\left|A\right| m /\delta)}.
\end{eqnarray*}
Then, with probability at least $1 - \delta/2$,  $Q^{mOIM}_t(x, a)
> Q^*(x,a)-\epsilon_1$ for all $t=1,2,\ldots$
\end{lem}

According to the previous lemma,
\begin{eqnarray} \label{e:oimlemma_1}
  \sum_y \hat P_t(x, a, y) \bigl( \hat R_t(x, a, y) + \gamma V^*(y) \bigr)
   - Q^*(x,a) \geq - \beta/ \sqrt{N_t(x,a)}
\end{eqnarray}
with probability $1-\delta/2$.

We will show that
\begin{eqnarray}  \label{e:oimlemma_2}
\frac{\rmax}{N_t(x,a)(1-\gamma)} + (1-\gamma)\epsilon_1 \geq
\frac{\beta}{\sqrt{N_t(x,a)}}.
\end{eqnarray}
For $N_t(x,a) \leq \frac{\rmax}{(1-\gamma)^2\epsilon_1}$, the
first term dominates the l.h.s. and we can omit the second term
(and prove the stricter inequality). In the following, we proceed
by a series of equivalent transformations:
\begin{eqnarray*}
\frac{\rmax}{N_t(x,a)(1-\gamma)} &\geq&
  \frac{\beta}{\sqrt{N_t(x,a)}}, \\
\frac{\rmax}{\beta(1-\gamma)} &\geq&
  \sqrt{N_t(x,a)}, \\
\frac{\rmax^2}{\beta^2(1-\gamma)^2} &\geq&
  N_t(x,a),
\end{eqnarray*}
which is implied by the stricter inequality
\begin{eqnarray*}
\frac{\rmax^2}{\beta^2(1-\gamma)^2} &\geq&
  \frac{\rmax}{(1-\gamma)^2\epsilon_1}, \\
\rmax &\geq&
  \frac{\beta^2}{\epsilon_1},
\end{eqnarray*}
which holds by the assumption of the lemma. If the relation is
reversed, then the first term can be omitted, leading to
\begin{eqnarray*}
(1-\gamma)\epsilon_1 &\geq&
  \frac{\beta}{\sqrt{N_t(x,a)}}, \\
\frac{\beta}{(1-\gamma)\epsilon_1} &\leq&
  \sqrt{N_t(x,a)}, \\
\frac{\beta^2}{(1-\gamma)^2\epsilon_1^2} &\leq&
  N_t(x,a), \\
\end{eqnarray*}
which is implied by the stricter inequality
\begin{eqnarray*}
\frac{\beta^2}{(1-\gamma)^2\epsilon_1^2} &\leq&
  \frac{\rmax}{(1-\gamma)^2\epsilon_1}, \\
\rmax &\geq&
  \frac{\beta^2}{\epsilon_1},
\end{eqnarray*}
similarly to the previous case.

At step $t$, a number of DP updates are carried out. We proceed by
induction on the number of DP-updates. Initially, $Q^{(0)}(x,a)
\geq Q^*(x,a) - \epsilon_1$, then
\begin{eqnarray*}
Q^{(i+1)} (x,a) &=&
  \sum_{y} \hat P_t(x, a, y) \bigl( \hat R_t(x, a, y) + \gamma V^{(i)}(y)
    \bigr) + \frac{\vo}{N_t(x,a)} \\
   &\geq& \sum_{y} \hat P_t(x, a, y) \bigl( \hat R_t(x, a, y) + \gamma
    (V^*(y)-\epsilon_1)
    \bigr) + \frac{\vo}{N_t(x,a)} \\
  &\geq& Q^*(x,a) - \beta/ \sqrt{N_t(x,a)} -\gamma\epsilon_1 + \frac{\vo}{N_t(x,a)} \\
  &\geq& Q^*(x,a) -\gamma\epsilon_1 - (1-\gamma)\epsilon_1 = Q^*(x,a) -\epsilon_1,
\end{eqnarray*}
where we applied  (\ref{e:oimlemma_1}), (\ref{e:oimlemma_2}) and
the induction assumption.
  \eproof

Define the $H$-step truncated value function of policy $\pi$ as
$Q^\pi(x,a,H) := E\Bigl( \sum_{t=0}^H \gamma^t r_t \Bigm|
x\!=\!x_0, a\!=\!a_0 \Bigr)$.

\begin{lem}[{[KS]} Lemma 2] \label{lem:truncation}
 Let $\epsilon>0$ and consider an MDP
$M=(X,A,P,R,\gamma)$. If
\[
 H \geq  \frac{1}{1-\gamma}\log \frac{\rmax^0}{\epsilon(1-\gamma)} ,
\]
then
\[
  Q^\pi(x,a,H) \leq Q^\pi(x,a) \leq Q^\pi(x,a,H) + \epsilon
\]
for any $(x,a) \in X\times A$.
\end{lem}
 \bproof
Let $\Xi(x,a)$ denote the set of infinite trajectories starting in
$(x,a)$, and for any trajectory $\xi \in \Xi(x,a)$, let $\xi_H$
denote its $H$-step truncation. Furthermore, denote the discounted
total reward along a trajectory $\xi$ by $v(\xi)$. Clearly,
\begin{eqnarray*}
  Q^\pi(x,a) &=& E_\xi [v(\xi)] = \sum_{\xi \in \Xi(x,a)} \Pr(\xi)
  v(\xi) \qquad\textrm{and} \\
  Q^\pi(x,a,H) &=& E_\xi [v(\xi_H)] = \sum_{\xi \in \Xi(x,a)} \Pr(\xi)
  v(\xi_H).
\end{eqnarray*}
Fix a trajectory $p$, along which the agent receives rewards $r_1,
r_2, \ldots$, for which
\begin{eqnarray*}
 v(\xi_H) &=& \sum_{t=0}^{H-1} \gamma^{t} r_{t+1}  \qquad\textrm{and} \\
 v(\xi) &=& \sum_{t=0}^\infty \gamma^{t} r_{t+1} = v(\xi_H) + \sum_{t=H}^\infty \gamma^{t}
 r_{t+1}.
\end{eqnarray*}
It is trivial that $v(\xi) \geq v(\xi_H)$, as the additional terms
are all nonnegative by assumption. On the other hand,
\begin{eqnarray*}
 \sum_{t=H}^\infty \gamma^{t} r_{t+1} &\leq&
   \sum_{t=H}^\infty \gamma^{t} \rmax^0 =
   \frac{\gamma^H}{1-\gamma} \rmax^0,
\end{eqnarray*}
which is smaller than $\epsilon$ if $H \geq \log
\frac{\epsilon(1-\gamma)}{\rmax^0} / \log \gamma$ (which follows
from the assumption of the lemma and the inequality $-\log \gamma
> 1-\gamma$), that is,
\[
  v(\xi) \leq v(\xi_H) + \epsilon.
\]
As the relations hold for each trajectory in $\Xi(x,a)$, they hold
for the expected value, too.
 \eproof

The following lemma tells that \ourmethod\ and its modified
version learn almost the same values with high probability.
\begin{lem} \label{lem:mOIMlemma}
For any $\epsilon>0$, $\delta>0$,
\begin{eqnarray*}
  \epsilon' &:=& \frac{(1-\gamma)^2}{\xsize \left( 1-\gamma + \rmax^0
\right)}\cdot \epsilon, \\
  m &\geq& \frac{2\max\{1,\rmax^0\}^2}{{\epsilon'}^2} \ln \frac{2}{\delta} ,
\end{eqnarray*}
for any MDP $M$ and any $(x,a)\in X\times A$,
\[
  \left| Q_M^{\pi^{m\ourmethod}}(x,a) - Q_M^{\pi^{\ourmethod}}(x,a)
  \right| \leq 2\epsilon
\]
with probability at least $1-2\delta$.
\end{lem}
 \bproof
The model estimates of the two algorithm-variants are identical on
not-yet-known states where the visit count is less than $m$. On
known pairs, we can apply Lemma~\ref{lem:knowingPR} to both
model-estimates to see that they are $\epsilon'$-close to the true
model parameters with probability at least $1-\delta$.
Consequently, they are $2\epsilon'$-close to each other with at
least $1-2\delta$ probability. Applying Lemma \ref{lem:simulation}
proves the statement of the lemma.
 \eproof

\begin{lem}[Lemma 3 of {[SL]}] \label{lem:explorationevent}
Let $M = (X,A,P,R,\gamma)$ be an MDP, $K$ a set of state-action
pairs, $\bar M$ an MDP equal to $M$ on $K$ (identical transition
and reward functions), $\pi$ a policy, and $H$ some positive
integer. Let $A_M$ be the event that a state-action pair not in
$K$ is encountered in a trial generated by starting from $(x,a)$
and following $\pi$ for $H$ steps in $M$. Then,
\begin{equation}
  Q^\pi_M(x,a) \geq Q^\pi_{\bar M}(x,a) - \frac{\rmax^0}{1-\gamma} \Pr( A_M ).
\end{equation}
\end{lem}
 \bproof
Let $\Xi$ be the set of $H$-step long trajectories, and let $\Xi^K
\subset \Xi$ be the set of trajectories for which all occurring
$(x_t,a_t)$ pairs are in $K$. For any $\xi \in \Xi$, let
$\Pr_M(\xi)$ denote the probability of that trajectory happening
in MDP $M$.

Let $v(\xi)$ be the discounted total reward received by the agent
along the $H$-step trajectory $\xi \in \Xi$. Now, we have the
following:
\begin{eqnarray*}
 Q^\pi_{\bar M}(x,a) &=&  \sum_{\xi\in\Xi} \Pr{}_{\bar M}(\xi) v(\xi) \\
   &=& \sum_{\xi\in\Xi^K} \Pr{}_{\bar M}(\xi) v(\xi)
    +\sum_{\xi\in\Xi \backslash \Xi^K} \Pr{}_{\bar M}(\xi) v(\xi) \\
   &\leq& \sum_{\xi\in\Xi^K} \Pr{}_{\bar M}(\xi) v(\xi)
    + \sum_{\xi\in\Xi \backslash \Xi^K} \Pr{}_{\bar M}(\xi)
    \frac{\rmax^0}{1-\gamma} \\
   &\leq& \sum_{\xi\in\Xi^K} \Pr{}_{\bar M}(\xi) v(\xi)
    + \Pr(A_{\bar{M}})
    \frac{\rmax^0}{1-\gamma} \\
   &=& \sum_{\xi\in\Xi^K} \Pr{}_{ M}(\xi) v(\xi)
    + \Pr(A_{M}) \frac{\rmax^0}{1-\gamma} \\
   &\leq& Q_M^\pi(x,a)
    + \Pr(A_{M}) \frac{\rmax^0}{1-\gamma}.
\end{eqnarray*}
 \eproof

\begin{thm}
For any $\epsilon>0, \, \delta>0$, let
\begin{eqnarray*}
\epsilon_1 &:=& \epsilon/6 \\
  \epsilon_2 &:=&  \frac{(1-\gamma)^2}{\xsize \left( 1-\gamma +
  \rmax^0
\right)}\cdot \epsilon_1, \\
H &:=& \frac{1}{1-\gamma} \ln \frac{\rmax^0}{\epsilon_1
(1-\gamma)} \\
  m &:=& \frac{2\max\{1,\rmax^0\}^2}{\epsilon_2^2} \ln
\frac{8}{\delta}.
\end{eqnarray*}
OIM converges almost surely to a near-optimal policy in polynomial
time if started with
\[
  \rmax = \frac{2(\rmax^0)^2\ln(2\left|X\right|
        \left|A\right| m /\delta)}{\epsilon_1
        (1-\gamma)^3},
\]
that is, with probability $1-\delta$, the number of timesteps
where $Q^{\pi^\ourmethod}(x_t,a_t) > Q^*(x_t,a_t)-\epsilon$ does
not hold, is at most
\[
\frac{2m \left|X\right| \left|A\right| H
\rmax^0}{\epsilon_1(1-\gamma)} \ln \frac{4}{\delta}
\]
\end{thm}

\begin{rem}
When expressed in terms of MDP parameters, time requirement is
\begin{eqnarray*}
&&\frac{864 \left|X\right|^3 \left|A\right| \rmax^0
\max\{1,\rmax^0\}^2 ( 1-\gamma + \rmax^0 )^2
}{\epsilon^3(1-\gamma)^4} \ln \frac{6 \rmax^0}{\epsilon
(1-\gamma)} \ln \frac{4}{\delta} \ln \frac{8}{\delta} \\
&&= O\left(
 \frac{\left|X\right|^3 \left|A\right| (\rmax^0)^5
}{\epsilon^3(1-\gamma)^4} \ln \frac{ \rmax^0}{\epsilon (1-\gamma)}
\ln^2 \frac{1}{\delta}
 \right)
\end{eqnarray*}
and the required initialization value is
\begin{eqnarray*}
  \rmax &=& \frac{12(\rmax^0)^2}{\epsilon
        (1-\gamma)^3} \ln \left(\frac{12\left|X\right|^3
        \left|A\right| \max\{1,\rmax^0\}^2 ( 1-\gamma + \rmax^0 )^2 }{\delta\epsilon(1-\gamma)^2}
        \ln\frac{8}{\delta}\right) \\
        &=& O\left(
\frac{(\rmax^0)^2}{\epsilon
        (1-\gamma)^3} \ln \left(\frac{\left|X\right|^3
        \left|A\right| (\rmax^0 )^4 }{\delta\epsilon(1-\gamma)^2}
        \ln\frac{1}{\delta}\right)
 \right)
\end{eqnarray*}
\end{rem}

 \bproof

Let $M$ denote the true (and unknown) MDP, let $\hat{M}$ be the
approximate model of \ourmethod. 

An $(x,a)$ pair is considered \emph{known} if it has been visited
at least $m$ times. According to Lemma~\ref{lem:knowingPR}, for a
known pair $(x,a)$, the model estimates $\hat P(x,a,\cdot)$ and
$\hat R(x,a,\cdot)$ are $\epsilon_2$-close to the true values with
probability at least $1-\delta/4$.

Define the MDP $\bar{M}$ so that it is identical to $M$ for known
pairs, and equals $\hat{M}$ for unknown pairs. The parameters of
$\hat{M}$ and $\bar{M}$ are identical on unknown pairs and
$\epsilon_2$-close for known pairs (with probability
$1-\delta/4$), so, by Lemma~\ref{lem:simulation},
\begin{equation} \label{e:oimlemma_3}
|Q^\pi_{\hat{M}}(x,a) - Q^\pi_{\bar{M}}(x,a) | < \epsilon_1
\end{equation}
for any policy $\pi$ and any $(x,a) \in X\times A$.

Let
\[
 H:= \frac{1}{1-\gamma} \ln \frac{\rmax^0}{\epsilon_1 (1-\gamma)}.
\]
By Lemma~\ref{lem:truncation},
\begin{equation} \label{e:thmXX}
 |Q^\pi_M(x,a,H) - Q^\pi_M(x,a)| < \epsilon_1
\end{equation}
holds for the $H$-step truncated value function for any $(x,a)$,
$\pi$.

Consider a state-action pair $(x_1,a_1)$ and a $H$-step long
trajectory generated by $\pi$. Let $K$ be the set of known $(x,a)$
pairs and let $A_M$ be the event that an unknown pair is
encountered along the trajectory. Then, by Lemma
\ref{lem:explorationevent},
\begin{equation} \label{e:oimlemma_4}
  Q^\pi_M(x_1,a_1) \geq Q^\pi_{\bar M}(x_1,a_1) - \frac{\rmax^0}{1-\gamma} \Pr( A_M ).
\end{equation}

By applying Lemma \ref{lem:mOIMlemma} to $\epsilon_1$, $\delta/4$,
we get that the above setting of $\rmax$ ensures that the original
and the modified version of \ourmethod\ behaves similarly:
\begin{equation} \label{e:thmXX2}
  \left| Q_M^{\pi^{m\ourmethod}}(x,a) - Q_M^{\pi^{\ourmethod}}(x,a)
  \right| \leq 2\epsilon_1
\end{equation}
with probability at least $1-\delta/2$. Furthermore, by
Lemma~\ref{lem:optimismpreserved} (with $\epsilon \leftarrow
\epsilon_1$ and $\delta \leftarrow \delta/4$), the modified
algorithm preserves the optimism of the value function with
probability at least $1-\delta/4$:
\[
 Q^{mOIM}_t(x, a) > Q^*(x,a)-\epsilon_1
\]

To conclude the proof, we separate two cases (following the line
of thoughts of Theorem 1 in [SL]). In the first case, an
exploration step will occur with high probability:  Suppose that
$\Pr(A_M) > \epsilon_1 (1-\gamma) / \rmax^0$, that is, an unknown
pair is visited in $H$ steps with high probability. This can
happen at most $m \left|X\right| \left|A\right| $ times, so by
Azuma's bound, with probability $1-\delta/4$, all $(x,a)$ will
become known after $\frac{2m \left|X\right| \left|A\right| H
\rmax^0}{\epsilon_1(1-\gamma)} \ln \frac{4}{\delta} $ exploration
steps.

On the other hand, if $\Pr(A_M) \leq \epsilon_1 (1-\gamma) /
\rmax^0$, then the policy is near-optimal with probability
$1-\delta$:
\begin{eqnarray*}
 && Q_M^{\pi^{\ourmethod}}(x_1,a_1) \geq Q_M^{\pi^{\ourmethod}}(x_1,a_1, H)  \\
 && \geq Q_{\bar M}^{\pi^{\ourmethod}}(x_1,a_1, H) - \frac{\rmax^0}{1-\gamma} \Pr( A_M )  \\
 && \geq Q_{\bar M}^{\pi^{\ourmethod}}(x_1,a_1, H) - \epsilon_1
   \geq Q_{\bar M}^{\pi^{\ourmethod}}(x_1,a_1) - 2 \epsilon_1  \\
 &&  \geq Q_{\hat M}^{\pi^{\ourmethod}}(x_1,a_1) - 3 \epsilon_1 \\
 &&  \geq Q_{\hat M}^{\pi^{m\ourmethod}}(x_1,a_1) - 5 \epsilon_1
 \\
 &&  \geq Q^*(x_1,a_1) - 6 \epsilon_1 \\
 &&  = Q^*(x_1,a_1) - \epsilon, \\
\end{eqnarray*}
where we applied (in this order) the property that truncation
decreases the value function; Eq.~(\ref{e:oimlemma_4}); our
assumption; Eq.~(\ref{e:thmXX}); Eq.~(\ref{e:oimlemma_3});
Eq.~(\ref{e:thmXX2}); Lemma \ref{lem:optimismpreserved} and the
definition of $\epsilon_1$.

 \eproof

\section{A dimension-respecting version of the proof}

For the proof, we shall follow the technique of
\citet{Kearns02Near-Optimal} and \citet{Strehl06Analysis}, and
will use the shorthands [KS] and [SL] for referring to them. We
will proceed by a series of lemmas.

\begin{lem} \label{lem:knowingPRb}
Consider an MDP $M =
(X,A,P,R,\gamma)$, and let $(x,a)$ be a state-action pair that has
been visited at least $m$ times. Let $\hat{P}(x,a,y)$ and
$\hat{R}(x,a,y)$ denote the obtained empirical estimates, let
$\epsilon>0$ and $\delta>0$ be arbitrary positive values. If
\[
 m \geq \frac{2}{\epsilon^2} \ln \frac{2}{\delta} ,
\]
then for all $y\in X$,
\begin{eqnarray*}
  \left| P(x,a,y) R(x,a,y) - \hat P(x,a,y)\hat R(x,a,y) \right| &\leq& \epsilon\rmax^0 \qquad\textrm{and} \\
  \left| P(x,a,y) - \hat P(x,a,y) \right| &\leq& \epsilon
\end{eqnarray*}
holds with probability at least $1-\delta$.
\end{lem}
 \bproof
The second statement is already proven, so let us consider the first one.
Suppose that $(x,a)$ is visited $k$ times at steps $t_1, \ldots,
t_k$. Define the random variables
\[
  W_i(y) = \left\{%
\begin{array}{ll}
    r_{t_i+1}, & \hbox{if $x_{t_i+1} = y$;} \\
    0, & \hbox{otherwise.} \\
\end{array}%
\right.
\]
In this case, $E[W_i(y)] = P(x,a,y) R(x,a,y)$,
$W_i(y)-P(x,a,y)R(x,a,y)$ is a martingale and is bounded by
$\rmax^0$ (note that we are considering only states $x,y\in X$, that is,
the garden-of-Eden state $x_E$ is excluded. Therefore, $\rmax^0$ is indeed
an upper bound on $R(x,a,y)$), so we can apply Azuma's lemma with $a = k\epsilon\rmax^0$ to
get
\[
  \Pr \left[ \left| \frac{1}{k}\sum_{i=1}^k W_i - P(x,a,y)R(x,a,y) \right| \geq \epsilon\rmax^0 \right] \leq 2\exp \left( -
  \frac{(\epsilon\rmax^0)^2 k}{2(\rmax^0)^2}\right)
  \leq 2\exp \left( -\frac{\epsilon^2 m}{2}\right).
\]
The right-hand side is less than $\delta$ for
\[
m \geq \frac{2}{\epsilon^2} \ln \frac{2}{\delta}.
\]
 \eproof

The following is a minor modification of {[KS]} lemma 4, and [SL]
Lemma 1. The result tells that if the parameters of two MDPs are
very close to each other, then the value functions in the two MDPs
will also be similar.

\begin{lem} \label{lem:simulationb}
Let $\epsilon>0$, and consider two MDPs $M = (X,A,P,R,\gamma)$ and
$\bar M = (X,A,\bar P,\bar R,\gamma)$ that differ only in their
transition and reward functions, furthermore, their difference is
bounded:
\begin{eqnarray*}
  \left| P(x,a,y)R(x,a,y) - \bar P(x,a,y)\bar R(x,a,y) \right| &\leq& \epsilon'\rmax^0 \qquad\textrm{and} \\
  \left| P(x,a,y) - \bar P(x,a,y) \right| &\leq& \epsilon'
\end{eqnarray*}
for all $(x,a,y)\in X\times A\times X$ and
\begin{eqnarray*}
  \epsilon' &:=& \frac{(1-\gamma)^2}{\xsize } \epsilon.
\end{eqnarray*}
Then for any policy $\pi$ and any $(x,a)\in X\times A$,
\[
  \left| Q^\pi(x,a) - \bar Q^\pi(x,a) \right| \leq \epsilon\rmax^0.
\]
\end{lem}
 \bproof
Let $\Delta := \max_{(x,a)\in X\times A} |Q^\pi(x,a) - \bar
Q^\pi(x,a)|$, and note that for any $x\in X$,
\begin{eqnarray*}
 |V^\pi(x) - \bar V^\pi(x)| &=& \left| \sum_a \pi(x,a) (Q^\pi(x,a) - \bar
Q^\pi(x,a)) \right| \leq \sum_a \pi(x,a) \Delta = \Delta
\end{eqnarray*}
For a fixed $(x,a)$ pair,
\begin{eqnarray*}
 && \Delta = \left| Q^\pi (x,a) - \bar Q^\pi(x,a) \right| \\
 && = \left| \sum_{y\in X} P(x,a,y) \Bigl(R(x,a,y) + \gamma V^\pi(y) \Bigr)
    - \sum_{y\in X} \bar P(x,a,y) \Bigl(\bar R(x,a,y) + \gamma \bar V^\pi(y) \Bigr) \right| \\
 && \leq \sum_{y\in X}  \left|P(x,a,y)R(x,a,y)-\bar P(x,a,y) \bar R(x,a,y)  \right| \\
 && \quad + \left| \sum_{y\in X} \Bigl[P(x,a,y)-\bar P(x,a,y)\Bigr] \Bigl(\gamma V^\pi(y) \Bigr) \right|
     + \left| \sum_{y\in X} \bar P(x,a,y) \Bigl( \gamma \Bigl[V^\pi(y) - \bar V^\pi(y)\Bigr] \Bigr) \right| \\
 && \leq \xsize \epsilon'\rmax^0 + \sum_{y\in X} \epsilon' \Bigl| \gamma V^\pi(y) \Bigr|
    +  \sum_{y\in X} \bar P(x,a,y) (\gamma \Delta ) \\
 && \leq  \xsize \epsilon'\rmax^0 + \xsize\epsilon' \frac{\gamma\rmax^0}{1-\gamma}
    +    \gamma \Delta.
\end{eqnarray*}
Therefore,
\[
\Delta \leq \frac{\xsize \epsilon'  \rmax^0}{(1-\gamma)^2} = \epsilon \rmax^0
\]
 \eproof

Let us introduce a modified version of \ourmethod\ that behaves
exactly like the old one, except that in each $(x,a)$ pairs, it
performs at most $m$ updates. If a pair is visited more than $m$
times, the modified algorithm leaves the counters unchanged.

The following result is a modification of {[SL]}'s Lemma 7.

\begin{lem} \label{lem:betaineqb}
Suppose that the modified \ourmethod\ (stopping after $m$ updates)
is executed on an MDP $M= (X,A,P,R,\gamma)$ with
\begin{eqnarray*}
 m &:=&
\frac{2}{\epsilon^2} \ln \frac{2}{\delta} , \\
  \beta &:=& \frac{\rmax^0}{1-\gamma} \sqrt{2\ln(2\left|X\right|
\left|A\right| m /\delta)}.
\end{eqnarray*}
Then, with probability at least $1 - \delta$,
\[
 Q^*(x,a) -
\sum_{y\in X} \hat{P}_t(x,a,y) \left[ \hat{R}_t (x,a,y) + \gamma
V^*(y) \right]
 \leq \beta/\sqrt{k}
\]
for all $t=1,2,\ldots$
\end{lem}
\bproof
The proof is identical to the proof of Lemma~\ref{lem:betaineq}.
\eproof

The following result shows that the modified \ourmethod\ algorithm
preserves the optimism of the value function with high
probability.

\begin{lem} \label{lem:optimismpreservedb}
Let $\epsilon_1>0$ and suppose that the modified \ourmethod\ is
executed on an MDP $M=(X,A,P,R,\gamma)$ with
\[
  \rmax \geq \frac{\beta^2}{\epsilon_1\rmax^0}
\]
where
\begin{eqnarray*}
 m &:=&
\frac{2}{\epsilon^2} \ln \frac{2}{\delta} , \\
  \beta &:=& \frac{\rmax^0}{1-\gamma} \sqrt{2\ln(2\left|X\right|
\left|A\right| m /\delta)}.
\end{eqnarray*}
Then, with probability at least $1 - \delta/2$,  $Q^{mOIM}_t(x, a)
> Q^*(x,a)-\epsilon_1\rmax^0$ for all $t=1,2,\ldots$
\end{lem}

According to the previous lemma,
\begin{eqnarray} \label{e:oimlemma_1b}
  \sum_y \hat P_t(x, a, y) \bigl( \hat R_t(x, a, y) + \gamma V^*(y) \bigr)
   - Q^*(x,a) \geq - \beta/ \sqrt{N_t(x,a)}
\end{eqnarray}
with probability $1-\delta/2$.

We will show that
\begin{eqnarray}  \label{e:oimlemma_2b}
\frac{\rmax}{N_t(x,a)(1-\gamma)} + (1-\gamma)\epsilon_1\rmax^0 \geq
\frac{\beta}{\sqrt{N_t(x,a)}}.
\end{eqnarray}
For $N_t(x,a) \leq \frac{\rmax}{\rmax^0}\frac{1}{(1-\gamma)^2\epsilon_1}$, the
first term dominates the l.h.s. and we can omit the second term
(and prove the stricter inequality). In the following, we proceed
by a series of equivalent transformations:
\begin{eqnarray*}
\frac{\rmax}{N_t(x,a)(1-\gamma)} &\geq&
  \frac{\beta}{\sqrt{N_t(x,a)}}, \\
\frac{\rmax}{\beta(1-\gamma)} &\geq&
  \sqrt{N_t(x,a)}, \\
\frac{\rmax^2}{\beta^2(1-\gamma)^2} &\geq&
  N_t(x,a),
\end{eqnarray*}
which is implied by the stricter inequality
\begin{eqnarray*}
\frac{\rmax^2}{\beta^2(1-\gamma)^2} &\geq&
  \frac{\rmax}{\rmax^0}\frac{1}{(1-\gamma)^2\epsilon_1}, \\
\rmax &\geq&
  \frac{\beta^2}{\epsilon_1\rmax^0},
\end{eqnarray*}
which holds by the assumption of the lemma. If the relation is
reversed, then the first term can be omitted, leading to
\begin{eqnarray*}
(1-\gamma)\epsilon_1\rmax^0 &\geq&
  \frac{\beta}{\sqrt{N_t(x,a)}}, \\
\frac{\beta}{(1-\gamma)\epsilon_1\rmax^0} &\leq&
  \sqrt{N_t(x,a)}, \\
\frac{\beta^2}{(1-\gamma)^2\epsilon_1^2(\rmax^0)^2} &\leq&
  N_t(x,a), \\
\end{eqnarray*}
which is implied by the stricter inequality
\begin{eqnarray*}
\frac{\beta^2}{(1-\gamma)^2\epsilon_1^2(\rmax^0)^2} &\leq&
  \frac{\rmax}{\rmax^0}\frac{1}{(1-\gamma)^2\epsilon_1}, \\
\rmax &\geq&
  \frac{\beta^2}{\epsilon_1\rmax^0},
\end{eqnarray*}
similarly to the previous case.

At step $t$, a number of DP updates are carried out. We proceed by
induction on the number of DP-updates. Initially, $Q^{(0)}(x,a)
\geq Q^*(x,a) - \epsilon_1\rmax^0$, then
\begin{eqnarray*}
Q^{(i+1)} (x,a) &=&
  \sum_{y} \hat P_t(x, a, y) \bigl( \hat R_t(x, a, y) + \gamma V^{(i)}(y)
    \bigr) + \frac{\vo}{N_t(x,a)} \\
   &\geq& \sum_{y} \hat P_t(x, a, y) \bigl( \hat R_t(x, a, y) + \gamma
    (V^*(y)-\epsilon_1\rmax^0)
    \bigr) + \frac{\vo}{N_t(x,a)} \\
  &\geq& Q^*(x,a) - \beta/ \sqrt{N_t(x,a)} -\gamma\epsilon_1\rmax^0 + \frac{\vo}{N_t(x,a)} \\
  &\geq& Q^*(x,a) -\gamma\epsilon_1\rmax^0 - (1-\gamma)\epsilon_1\rmax^0 = Q^*(x,a) -\epsilon_1\rmax^0,
\end{eqnarray*}
where we applied  (\ref{e:oimlemma_1}), (\ref{e:oimlemma_2}) and
the induction assumption.
  \eproof

Define the $H$-step truncated value function of policy $\pi$ as
$Q^\pi(x,a,H) := E\Bigl( \sum_{t=0}^H \gamma^t r_t \Bigm|
x\!=\!x_0, a\!=\!a_0 \Bigr)$.

\begin{lem} \label{lem:truncationb}
 Let $\epsilon>0$ and consider an MDP
$M=(X,A,P,R,\gamma)$. If
\[
 H \geq  \frac{1}{1-\gamma}\log \frac{1}{\epsilon(1-\gamma)} ,
\]
then
\[
  Q^\pi(x,a,H) \leq Q^\pi(x,a) \leq Q^\pi(x,a,H) + \epsilon \rmax^0
\]
for any $(x,a) \in X\times A$.
\end{lem}
 \bproof
Let $\Xi(x,a)$ denote the set of infinite trajectories starting in
$(x,a)$, and for any trajectory $\xi \in \Xi(x,a)$, let $\xi_H$
denote its $H$-step truncation. Furthermore, denote the discounted
total reward along a trajectory $\xi$ by $v(\xi)$. Clearly,
\begin{eqnarray*}
  Q^\pi(x,a) &=& E_\xi [v(\xi)] = \sum_{\xi \in \Xi(x,a)} \Pr(\xi)
  v(\xi) \qquad\textrm{and} \\
  Q^\pi(x,a,H) &=& E_\xi [v(\xi_H)] = \sum_{\xi \in \Xi(x,a)} \Pr(\xi)
  v(\xi_H).
\end{eqnarray*}
Fix a trajectory $p$, along which the agent receives rewards $r_1,
r_2, \ldots$, for which
\begin{eqnarray*}
 v(\xi_H) &=& \sum_{t=0}^{H-1} \gamma^{t} r_{t+1}  \qquad\textrm{and} \\
 v(\xi) &=& \sum_{t=0}^\infty \gamma^{t} r_{t+1} = v(\xi_H) + \sum_{t=H}^\infty \gamma^{t}
 r_{t+1}.
\end{eqnarray*}
It is trivial that $v(\xi) \geq v(\xi_H)$, as the additional terms
are all nonnegative by assumption. On the other hand,
\begin{eqnarray*}
 \sum_{t=H}^\infty \gamma^{t} r_{t+1} &\leq&
   \sum_{t=H}^\infty \gamma^{t} \rmax^0 =
   \frac{\gamma^H}{1-\gamma} \rmax^0,
\end{eqnarray*}
which is smaller than $\epsilon\rmax^0$ if $H \geq \log
\epsilon(1-\gamma)/ \log \gamma$ (which follows
from the assumption of the lemma and the inequality $-\log \gamma
> 1-\gamma$), that is,
\[
  v(\xi) \leq v(\xi_H) + \epsilon \rmax^0.
\]
As the relations hold for each trajectory in $\Xi(x,a)$, they hold
for the expected value, too.
 \eproof

The following lemma tells that \ourmethod\ and its modified
version learn almost the same values with high probability.
\begin{lem} \label{lem:mOIMlemmab}
For any $\epsilon>0$, $\delta>0$,
\begin{eqnarray*}
  \epsilon' &:=& \frac{(1-\gamma)^2}{\xsize } \epsilon, \\
  m &\geq& \frac{2}{\epsilon^2} \ln \frac{2}{\delta} , \\
\end{eqnarray*}
for any MDP $M$ and any $(x,a)\in X\times A$,
\[
  \left| Q_M^{\pi^{m\ourmethod}}(x,a) - Q_M^{\pi^{\ourmethod}}(x,a)
  \right| \leq 2\epsilon \rmax^0
\]
with probability at least $1-2\delta$.
\end{lem}
 \bproof
The model estimates of the two algorithm-variants are identical on
not-yet-known states where the visit count is less than $m$. On
known pairs, we can apply Lemma~\ref{lem:knowingPRb} to both
model-estimates to see that they are $\epsilon'$-close to the true
model parameters with probability at least $1-\delta$.
Consequently, they are $2\epsilon'$-close to each other with at
least $1-2\delta$ probability. Applying Lemma \ref{lem:simulationb}
proves the statement of the lemma.
 \eproof

\begin{lem} \label{lem:explorationeventb}
Let $M = (X,A,P,R,\gamma)$ be an MDP, $K$ a set of state-action
pairs, $\bar M$ an MDP equal to $M$ on $K$ (identical transition
and reward functions), $\pi$ a policy, and $H$ some positive
integer. Let $A_M$ be the event that a state-action pair not in
$K$ is encountered in a trial generated by starting from $(x,a)$
and following $\pi$ for $H$ steps in $M$. Then,
\begin{equation}
  Q^\pi_M(x,a) \geq Q^\pi_{\bar M}(x,a) - \frac{\rmax^0}{1-\gamma} \Pr( A_M ).
\end{equation}
\end{lem}
 \bproof
The lemma is identical to Lemma~\ref{lem:explorationevent}.
 \eproof

\begin{thm}
For any $\epsilon>0, \, \delta>0$, let
\begin{eqnarray*}
\epsilon_1 &:=& \epsilon/6 \\
  \epsilon_2 &:=&  \frac{(1-\gamma)^2}{\xsize }\cdot \epsilon_1, \\
H &:=& \frac{1}{1-\gamma} \ln \frac{1}{\epsilon_1
(1-\gamma)} \\
  m &:=& \frac{2}{\epsilon_2^2} \ln
\frac{8}{\delta}.
\end{eqnarray*}
OIM converges almost surely to a near-optimal policy in polynomial
time if started with
\[
  \rmax = \frac{2\rmax^0\ln(2\left|X\right|
        \left|A\right| m /\delta)}{\epsilon_1
        (1-\gamma)^2},
\]
that is, with probability $1-\delta$, the number of timesteps
where $Q^{\pi^\ourmethod}(x_t,a_t) > Q^*(x_t,a_t)-\epsilon \rmax^0$ does
not hold, is at most
\[
\frac{2m \left|X\right| \left|A\right| H
}{\epsilon_1(1-\gamma)} \ln \frac{4}{\delta}
\]
\end{thm}

\begin{rem}
When expressed in terms of MDP parameters, time requirement is
\begin{eqnarray*}
&&\frac{864 \left|X\right|^3 \left|A\right|
}{\epsilon^3(1-\gamma)^4} \ln \frac{6 }{\epsilon
(1-\gamma)} \ln \frac{4}{\delta} \ln \frac{8}{\delta} \\
&&= O\left(
 \frac{\left|X\right|^3 \left|A\right|
}{\epsilon^3(1-\gamma)^4} \ln \frac{ 1}{\epsilon (1-\gamma)}
\ln^2 \frac{1}{\delta}
 \right)
\end{eqnarray*}
and the required initialization value is
\begin{eqnarray*}
  \rmax &=& \frac{12\rmax^0}{\epsilon
        (1-\gamma)^2} \ln \left(\frac{144\left|X\right|^3
        \left|A\right|  }{\delta\epsilon(1-\gamma)^4}
        \ln\frac{8}{\delta}\right) \\
        &=& O\left(
\frac{\rmax^0}{\epsilon
        (1-\gamma)^2} \ln \left(\frac{\left|X\right|^3
        \left|A\right|  }{\delta\epsilon(1-\gamma)^4}
        \ln\frac{1}{\delta}\right)
 \right)
\end{eqnarray*}
\end{rem}

 \bproof

Let $M$ denote the true (and unknown) MDP, let $\hat{M}$ be the
approximate model of \ourmethod. 

An $(x,a)$ pair is considered \emph{known} if it has been visited
at least $m$ times. According to Lemma~\ref{lem:knowingPR}, for a
known pair $(x,a)$, the model estimates $\hat P(x,a,\cdot)$ and
$\hat P(x,a,\cdot) \hat R(x,a,\cdot)$ are $\epsilon_2$-close and $\epsilon_2\rmax^0$ to the true values with
probability at least $1-\delta/4$.

Define the MDP $\bar{M}$ so that it is identical to $M$ for known
pairs, and equals $\hat{M}$ for unknown pairs. The parameters of
$\hat{M}$ and $\bar{M}$ are identical on unknown pairs and
$\epsilon_2$-close for known pairs (with probability
$1-\delta/4$), so, by Lemma~\ref{lem:simulation},
\begin{equation} \label{e:oimlemma_3b}
|Q^\pi_{\hat{M}}(x,a) - Q^\pi_{\bar{M}}(x,a) | < \epsilon_1 \rmax^0
\end{equation}
for any policy $\pi$ and any $(x,a) \in X\times A$.

Let
\[
 H:= \frac{1}{1-\gamma} \ln \frac{1}{\epsilon_1 (1-\gamma)}.
\]
By Lemma~\ref{lem:truncationb},
\begin{equation} \label{e:thmXXb}
 |Q^\pi_M(x,a,H) - Q^\pi_M(x,a)| < \epsilon_1 \rmax^0
\end{equation}
holds for the $H$-step truncated value function for any $(x,a)$,
$\pi$.

Consider a state-action pair $(x_1,a_1)$ and a $H$-step long
trajectory generated by $\pi$. Let $K$ be the set of known $(x,a)$
pairs and let $A_M$ be the event that an unknown pair is
encountered along the trajectory. Then, by Lemma
\ref{lem:explorationeventb},
\begin{equation} \label{e:oimlemma_4b}
  Q^\pi_M(x_1,a_1) \geq Q^\pi_{\bar M}(x_1,a_1) - \frac{\rmax^0}{1-\gamma} \Pr( A_M ).
\end{equation}

By applying Lemma \ref{lem:mOIMlemmab} to $\epsilon_1$, $\delta/4$,
we get that the above setting of $\rmax$ ensures that the original
and the modified version of \ourmethod\ behaves similarly:
\begin{equation} \label{e:thmXX2b}
  \left| Q_M^{\pi^{m\ourmethod}}(x,a) - Q_M^{\pi^{\ourmethod}}(x,a)
  \right| \leq 2\epsilon_1 \rmax^0
\end{equation}
with probability at least $1-\delta/2$. Furthermore, by
Lemma~\ref{lem:optimismpreservedb} (with $\epsilon \leftarrow
\epsilon_1$ and $\delta \leftarrow \delta/4$), the modified
algorithm preserves the optimism of the value function with
probability at least $1-\delta/4$:
\[
 Q^{mOIM}_t(x, a) > Q^*(x,a)-\epsilon_1 \rmax^0
\]

To conclude the proof, we separate two cases (following the line
of thoughts of Theorem 1 in [SL]). In the first case, an
exploration step will occur with high probability:  Suppose that
$\Pr(A_M) > \epsilon_1 \rmax^0(1-\gamma) $, that is, an unknown
pair is visited in $H$ steps with high probability. This can
happen at most $m \left|X\right| \left|A\right| $ times, so by
Azuma's bound, with probability $1-\delta/4$, all $(x,a)$ will
become known after $\frac{2m \left|X\right| \left|A\right| H
}{\epsilon_1(1-\gamma)} \ln \frac{4}{\delta} $ exploration
steps.

On the other hand, if $\Pr(A_M) \leq \epsilon_1 (1-\gamma)\rmax^0$, then the policy is near-optimal with probability
$1-\delta$:
\begin{eqnarray*}
 && Q_M^{\pi^{\ourmethod}}(x_1,a_1) \geq Q_M^{\pi^{\ourmethod}}(x_1,a_1, H)  \\
 && \geq Q_{\bar M}^{\pi^{\ourmethod}}(x_1,a_1, H) - \frac{\rmax^0}{1-\gamma} \Pr( A_M )  \\
 && \geq Q_{\bar M}^{\pi^{\ourmethod}}(x_1,a_1, H) - \epsilon_1 \rmax^0
   \geq Q_{\bar M}^{\pi^{\ourmethod}}(x_1,a_1) - 2 \epsilon_1 \rmax^0 \\
 &&  \geq Q_{\hat M}^{\pi^{\ourmethod}}(x_1,a_1) - 3 \epsilon_1 \rmax^0\\
 &&  \geq Q_{\hat M}^{\pi^{m\ourmethod}}(x_1,a_1) - 5 \epsilon_1 \rmax^0
 \\
 &&  \geq Q^*(x_1,a_1) - 6 \epsilon_1 \rmax^0 \\
 &&  = Q^*(x_1,a_1) - \epsilon \rmax^0, \\
\end{eqnarray*}
where we applied (in this order) the property that truncation
decreases the value function; Eq.~(\ref{e:oimlemma_4b}); our
assumption; Eq.~(\ref{e:thmXXb}); Eq.~(\ref{e:oimlemma_3b});
Eq.~(\ref{e:thmXX2b}); Lemma \ref{lem:optimismpreservedb} and the
definition of $\epsilon_1$.

 \eproof

\bibliography{rl}
\bibliographystyle{mlapa}

\end{document}